\documentclass{article} 
\usepackage{iclr2016_conference,times}
\usepackage{hyperref}
\usepackage{url}
\usepackage{amsthm}
\usepackage{amsmath}
\usepackage{graphicx}

\title{Manifold Regularized Discriminative Neural Networks}

\author{Shuangfei Zhai, Zhongfei (Mark) Zhang  \\
Department of Computer Science\\
Binghamton University\\
Vestal, New York 13902, USA \\
\texttt{szhai2@binghamton.edu, zhongfei@cs.binghamton.edu} \\
}

\begin{document}

\maketitle

\begin{abstract}
Unregularized deep neural networks (DNNs) can be easily overfit with a limited sample size. We argue that this is mostly due to the disriminative nature of DNNs which directly model the conditional probability (or score) of labels given the input. The ignorance of input distribution makes DNNs difficult to generalize to unseen data. Recent advances in regularization techniques, such as pretraining and dropout, indicate that modeling input data distribution (either explicitly or implicitly) greatly improves the generalization ability of a DNN. In this work, we explore the manifold hypothesis which assumes that instances within the same class lie in a smooth manifold. We accordingly
propose two simple regularizers to a standard discriminative DNN. The first one, named Label-Aware Manifold Regularization, assumes the availability of labels and penalizes large norms of the loss function w.r.t. data points. The second one, named Label-Independent Manifold Regularization, does not use label information and instead penalizes the Frobenius norm of the Jacobian matrix of prediction scores w.r.t. data points, which makes semi-supervised learning possible. We perform extensive control experiments on fully supervised and semi-supervised tasks using the MNIST, CIFAR10 and SVHN datasets and achieve excellent results. 
\end{abstract}

\section{Introduction}
Discriminative models are widely used for supervised machine learning tasks. In a probabilistic setting, they directly model the conditional probability of labels given the input $P(y|x)$ without considering the data distribution $P(x)$. Feed-forward neural networks are discriminative in nature, with which one can easily parameterize the conditional probability $P(y|x)$. However, it is well known that unregularized DNNs trained in this fashion (with MLE, for instance) suffers from serious overfitting problems with a limited sample size. 

In the past decade, we have witnessed several effective regularization techniques that help DNNs better generalize to unobserved data, most noticeably pretraining \citet{rbm,greedy} and dropout \citet{dropout}. In pretraining, several layers of either RBMs or regularized autoencoders are trained on unlabeled data. One then uses the weights of the stacked deep RBM/autoencoder as the initialization weights for a discriminative DNN, which is further trained on labeled dataset by adding an appropriate classifier layer on top. It is shown that pretrained weights perform very well as an implicit regularization as they prevent the model to get stuck in poor local optimums. The reason for its success is that with the pretrained RMB/autoencoder, the disriminative model gains (implicit) prior knowledge about the data distribution $P(x)$, thus is able to generalize better to unobserved data. Dropout works by randomly masking out part of both the input and hidden units during the discriminative training with SGD. \cite{dropout_augment} shows that the effectiveness of dropout can be understood by considering it as a kind of data augmentation \cite{augment}. From this perspective, dropout can be considered as implicitly modeling the data manifold by sampling from it.

Another motivation of our work is the recent findings of \cite{fooled}, where the authors show that the prediction of a trained discriminative DNN could be greatly changed by adding to the input a certain small perturbation that is indistinguishable by human eyes. This shows that even the state-of-the-art discriminative DNNs fail to generalize to what seems to be trivial examples. As a remedy, \cite{harness} proposes a way to generate adversarial examples (neighbors of sample data points that increase the loss function significantly) as the additional training data. 

In this paper, we propose two simple regularizers to the standard discriminative DNNs: Label-Aware Manifold Regularization (LAMR) and Label-Independent Manifold Regularization (LIMR). Our assumption is that data points within each class should lie in a smooth manifold. If we move from a data point $x$ by a tiny step $\epsilon$, our prediction should remain unchanged, as $x + \epsilon$ is very likely to still lie in the same manifold. In the label-aware version, we encourage the loss function to be flat around observations, which amounts to minimizing the norm of the loss function w.r.t. data points. In the label-independent version, we instead encourage the prediction scores to be flat, which can be computed without lablels. We then accordingly minimize the Frobenius norm of the Jacobian matrix of the prediction socres w.r.t. inputs. In addition, we propose stochastic approximations of the regularizers, which avoids explicitly computing the gradients and Jacobians. We perform extensive experiments on both fully labeled and semi-supervised versions of the MNIST dataset, and set state-of-the-art results.
\section{Model}
Consider a multi-way classification problem with data $\{(x_1, y_1), ..., (x_{N^l}, y_{N^l})\}$, where $x_i \in R^d$ and $y_i \in \{1, ..., K\}$ are the observation and label for the $i$-th instance, respectively; $N^l$ is the number of labeled instances. We also assume a set of $N^u$ unlabeled instances $\{x^u_1, ..., x^u_{N^u}\} $ available. A discriminative model minimizes an objective function in the following form:
\begin{equation}
\label{eq:discriminative}
 \underbrace{\frac{1}{N^l}\sum_{i=1}^{N^l} \ell(y_i, f(x^l_i;\theta))}_{J(\theta)} + R(\theta),
\end{equation}
where $f(x^l_i) \in R^K$ computes the predicted score of $x^l_i$ belonging to each of the $K$ class; $\ell$ is a loss function indicating the compatibility between the ground truth $y_i$ and the prediction $f(x^l_i)$; $R(\theta)$ is the regularization term. For example, $f(x;\theta)$ could be the conditional probability $p(y|x; \theta)$ represented by a DNN with a softmax layer, and $\ell(\cdot)$ is accordingly the cross entropy. Alternatively, as in \cite{dlsvm}, $f(x;\theta)$ could be the output of a DNN with a linear output layer, and $\ell$ is the one-vs-rest squared hinge loss. We use the later form for all the models in this paper. The focus of this paper is on $R(\theta)$, and we now propose two versions of regularization depending upon the availability of labels.

\subsection{Label-Aware Manifold Regularization}
In this version, we consider the fully supervised setting where all data points are labeled. According to the manifold assumption, data points within each class lie in a smooth manifold. Without any prior knowledge about the shape of the data manifold, we assume that moving any observation by a tiny step should still keep it in the same manifold. As a result, the loss function to be flat around each observation. In other words, the derivative of the loss function w.r.t. to input $\nabla_x(\ell(y, f(x;\theta)))$ should be small. This directly leads us to a regularization in the form:
\begin{equation}
\label{eq:reg1}
R^1(\theta) = \frac{\lambda }{N^l}\sum_{i=1}^{N^l} \|\nabla_{x^l_i}{\ell(y_i, f(x^l_i;\theta))}\|_2^2,
\end{equation}
where $\lambda > 0$ is the regularization strength factor. We call $R^1(\theta)$ Label-Aware Manifold Regularization (LAMR).

\subsection{Label-Independent Manifold Regularization}
LAMR assumes the availability of fully labeled data, which prohibits its application to semi-supervised settings where large portions of data are unlabeled. To this end, we propose a slightly different view on the regularization. The idea is that we can consider $f(x;\theta)$ as a mapping from the data manifold to the label manifold embedded in $R^K$. Following a similar reasoning, we then encourage $f(x;\theta)$, instead of the loss function, to be flat when we move the input along the data manifold. While the very same idea has been proposed in \cite{dcn} as deep contractive network (DCN), we further extend it by observing that this regularization also applies to unlabeled data. This makes this view particularly interesting as we are now able to better explore the data distribution with the help of unlabeled data. As a result, we minimize the Frobenius norm of the Jacobian matrix as follows:
\begin{equation}
\label{eq:reg2}
R^2(\theta) = \frac{\lambda}{N^l}\sum_{i=1}^{N^l}\|\nabla_{x^l_i}  f(x^l_i;\theta)\|_2^2 + \frac{\beta}{N^u}\sum_{i=1}^{N^u}\|\nabla_{x^u_i} f(x^u_i;\theta)\|_2^2.
\end{equation}
Here we have slightly overloaded the notation by letting $\nabla_{x} f(x;\theta)$ denote the Jacobian matrix of $f(x;\theta)$ w.r.t. $x$. We also allow different regularization strengths $\lambda$ and $\beta$ on labeled and unlabeled data.
We call $R^2(\theta)$ Label-Independent Manifold Regularization (LIMR). Note that LIMR is, different from LAMR, also loss independent. To see this, consider $f(x_i;\theta)$ as the output of a DNN with a linear output layer, LIMR would be the same no matter if we use the squared loss or hinge loss, but LAMR would be different for the two cases. This difference is verified in our experiments where we observe that LAMR achieves slightly lower test error on the fully supervised task.
On the other hand, LAMR and LIMR are also closely related, simply by noting that minimizing LIMR implicitly minimizes LAMR, as $R^1(\theta) \rightarrow 0$ when $R^2(\theta) \rightarrow 0$. 

\subsection{Stochastic Approximation}
Although it is possible to express $R^1(\theta)$ and $R^2(\theta)$ in a closed form for feed-forward neural networks, the resulting objective function could be complicated for DNNs and structured architectures such as CNNs. This makes computing the gradient w.r.t. model parameters inefficient. To address this issue, we take a simple stochastic approximation of the regularization terms. Let $g: R^d \mapsto R$ be a differentiable function, and $\epsilon \in \mathcal{N}(0, \sigma^2)$ be a draw from an isotropic Gaussian distribution with variance $\sigma^2$; then
\begin{equation}
g(x + \epsilon) \approx g(x) + \epsilon^T \nabla_x g(x),
\end{equation}
according to first order Taylor expansion. As a result, we have:
\begin{equation}
\mathrm{E}_{\epsilon}[g(x + \epsilon) - g(x)]^2 \approx \mathrm{E}_{\epsilon}[\epsilon^T \nabla_x g(x)]^2 = \sigma^2 \|\nabla_x g(x)\|_2^2,
\end{equation}
where in the last step we have marginalized out the isotropic Gaussian random variable. With this trick, we approximate the regularization terms for LAMR and LIMR as follows:
\begin{equation}
\label{eq:approximate}
\begin{split}
\tilde{R}^1(\theta) = &\frac{\lambda}{N^l}\sum_{i=1}^{N^l} \mathrm{E}_{\epsilon}[\ell(y_i, f(x^l_i + \epsilon;\theta)) - \ell(y_i, f(x^l_i;\theta))]^2 \\
\tilde{R}^2(\theta) = &\frac{\lambda}{N^l} \sum_{i=1}^{N^l} \mathrm{E}_{\epsilon}\|f(x^l_i + \epsilon;\theta) - f(x^l_i; \theta)\|_2^2 + \frac{\beta}{N^u} \sum_{i=1}^{N^u} \mathrm{E}_{\epsilon}\|f(x^u_i + \epsilon;\theta) - f(x^u_i; \theta)\|_2^2,
\end{split}
\end{equation}
where we have absorbed the scaling factor $\sigma^2$. In practice, we simulate the expectations in $\tilde{R}^1$ and $\tilde{R}^2$ by randomly sampling the Gaussian noise $\epsilon$ at each iteration of stochastic gradient descent (SGD), in the same way as \cite{dae}.

In theory, the stochastic approximation is more accurate when using a small $\sigma$. However, we find that it is beneficial to use a relatively large $\sigma$ in practice, which actually allows the regularization to explore points from the data manifold that are relatively far from the observations. 

\subsection{Connection With Existing Regularizers}
\textbf{Feature Noising} \cite{bishop}:One long existing trick for training neural networks is randomly adding noise to data during training. Marginally, its objective function can be expressed $\frac{1}{N^l}\sum_{i=1}^{N^l}\mathrm{E}_{\epsilon \in P_{noise}}{[\ell(y_i, f(x^l_i + \epsilon; \theta))]}$, where $P_{noise}$ is the additive noise distribution to generate samples around $x^l_i$. Feature noising thus implicitly explores the data manifold with $P_{noise}(x^l_i)$, in a way that is similar to LAMR. However, it does not stress the flatness of the loss function as LAMR does. To see this, consider the simplest case where $P_{noise} = \mathcal{N}(0, \sigma_{noise}^2I)$, we can then marginalize the noise by using the second order Taylor expansion as:
\begin{equation}
\label{eq:dropout}
\begin{split}
\underbrace{\frac{1}{N^l}\sum_{i=1}^{N^l}\ell(y_i, f(x^l_i; \theta))}_{J(\theta)} +  \underbrace{\frac{\sigma_{noise}^2}{N^l}\sum_{i=1}^{N^l} tr(H_i(\theta))}_{R(\theta)},
\end{split}
\end{equation}
where $H_i(\theta) = \nabla^2_{x^l_i} \ell(y_i, f(x^l_i;\theta)$ is the Hessian matrix of the loss function at $x^l_i$. We see that the regularization term now encourages the trace of Hessian matrix to be small around the observations, which corresponds to functions with low curvature. This is different from LAMR, as it is that the gradient is large under even though the trace of the Hessian matrix is small. Also, note that one variant of Dropout \cite{dropout} when mask out noise is only applied to the input layer can be analyzed in a similar way, and the readers are encouraged to see \cite{lrdrop} for detailed discussions.

\textbf{Manifold Tangent Classifier (MTC)} \cite{mtc}:
MTC is a semi-supervised classifier that considers the data manifold, which builds upon the Contractive Autoencoders (CAE) \cite{cae,caeh}. The idea is that one first trains a stacked CAE on all the data (with or without label) as in pretraining, from which the tangent bundles $B_x$ around each observation is then computed and stored.  
The corresponding regularization takes the form:
\begin{equation}
\label{eq:mtc}
R^{mtc}(\theta) = \frac{\lambda}{N^l}\sum_{i=1}^{N^l}\sum_{u_i \in B_{x^l}} [u_i^T \nabla_{x^l}\ell(y_i, x^l_i;\theta)]^2.
\end{equation}
In Equation \ref{eq:mtc}, the gradient of loss function is encouraged to be orthogonal to the tangent vectors $u_i$. The connection with LAMR can be observed if we replace $u_i$ with a random vector drawn from a Gaussian distribution $\mathcal{N}(0, \sigma^2I)$, and replace the inner sum with expectation over the random vector:
\begin{equation}
\frac{\lambda}{N^l}\sum_{i=1}^{N^l}\mathrm{E}_{u_i} [u_i^T \nabla_{x^l}\ell(y_i, x^l_i;\theta)]^2 =  \frac{\lambda \sigma^2}{N^l}\sum_{i=1}^{N^l}\|\nabla_{x^l}\ell(y_i, x^l_i;\theta)\|^2_2 \propto R^1(\theta).
\end{equation}
We see that forcing the gradient to be orthogonal to any random direction instead of the tangent vectors reduces MTC to LAMR. This highlights our assumption about the data manifold: if an observation is moved by a small random step, it is very likely to still fall into the same manifold. MTC on the other hand, explicitly characterizes the shape of the manifold with a dedicated model (CAE). As a result, LAMR puts a stronger regularization by enforcing $\nabla_{x^l}\ell(y_i, x^l_i;\theta)$ to be small, which also implicitly minimizes Equation \ref{eq:mtc}.
MTC is also able to perform semi-supervised learning, as unlabeled data could help better characterize the data manifold. LIMR takes a similar idea, but instead encourages the flatness of the prediction score function on the unlabeled data points. This also enables one to directly incorporate the regularization with the supervised objective, which makes the two-step pretraining fashion unnecessary.

\textbf{Deep Contractive Network (MTC)} \cite{dcn}: In the purely supervised setting, DCN is marginally equivalent to LIMR. However, there are two notable differences. First, we extend LIMR to the semi-supervised learning setting. Second, our stochastic approximation allows LIRM to be applied to deep neural networks with complicated structures (with convolutional layers, for example) easily. Moreover, as shown in the experiments, the stochastic approximation with a large noise allows us to explore a larger area of the input space, which puts more regularization on the flatness of the target function.

\textbf{Adversarial Training \cite{harness}:}
Recently in \cite{fooled}, the authors point out that the state-of-the-art DNNs can misclassify examples that are generated by adding to the training examples with certain small perturbations (adversarial examples). To address this issue, \cite{harness} propose a strategy to generate adversarial examples and add them to the training set along training. LAMR and LIMR are able to alleviate the very same problem from a different angle: reducing the chance that adversarial examples occur by restricting the loss function to be flat around the observations. 

\section{Experiments}
\subsection{Supervised Learning}
\label{sect:fc}
In the first set of experiments, we use the MNIST dataset to validate the regularization effects in the supervised learning setting. MNIST consists of 50000 training, 10000 validation and 10000 testing images of handwritten digits, each with shape 28x28. W train a three layer feed-forward neural network with size $500 \times 500 \times 500 \times 10$, where $10$ is the size of outputs. We use ReLU as the nonlinear activation function for each of the three hidden layers, and a linear output layer (whose output corresponds to $f(x_i;\theta)$) together with the one-vs-rest squared hinge loss as in \cite{dlsvm}. Parameters are initialized following \cite{xavier} and optimized with Adadelta \cite{adadelta} using a batch size of $100$. We set the standard deviation $\sigma$ as $0.5$, and vary the regularization strength $\lambda$ in Equation \ref{eq:approximate} for both LAMR and LIMR ($\beta=0$ in the fully supervised task), respectively. We show the results in Table \ref{tb:fc50k}. We see that without regularization ($\lambda = 0$), this three layer network achieves a test error rate of $1.64$. For both LAMR and LIMR, the error rate is significantly reduced and gradually changes as we vary the regularization strength. Overall, we see that the two regularizers achieve similar best results and are both pretty robust w.r.t. $\lambda$ within a certain range. 

\begin{table}[h]
\centering
\begin{tabular}{l | c  c c c c c c}
$\lambda$ for LAMR & 0 & 0.01 & 0.05 & 0.1 & 0.2 & 0.3 & 0.5 \\
 \hline
 validation error rate& 1.64 & 1.41 & 1.13 & \bf 1.01 & 1.11 & 1.04 & 1.08\\
 test error rate& 1.64 &1.4 &1.15 & \bf 0.95 &1.11 &0.99 &1.2 \\
 \vspace{1em}
\end{tabular}

\begin{tabular}{l | c c c c c c c}
$\lambda$ for LIMR & 0.01 & 0.05 & 0.1 & 0.3 & 0.7 & 1 & 1.5 \\
 \hline
 validation error rate& 1.39 &1.22 &0.95 &0.97& \bf 0.94 &0.94 &0.98\\
 test error rate& 1.45 &1.13 &0.96 &1.03 &\bf 0.96 &0.99 &1.01
\end{tabular}
\caption{The effect of varying regularization strength $\lambda$ for LAMR (top tabular) and LIMR (bottom tabular), no regularization is used when $\lambda = 0$. We see that both regularizers achieve significant better test errors (0.95 and 0.96) than the plain neural net without regularization (1.64).}
\label{tb:fc50k}
\end{table}

\begin{figure}
\begin{minipage}{0.5\textwidth}
\includegraphics[width=0.9\linewidth]{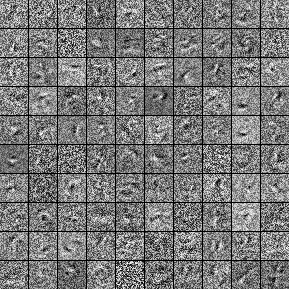}
\end{minipage}
\begin{minipage}{0.5\textwidth}
\includegraphics[width=0.9\linewidth]{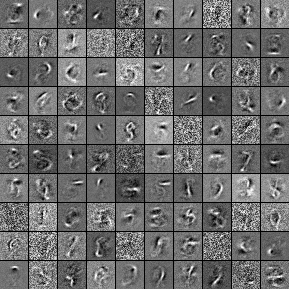}
\end{minipage}
\caption{Filters of the first layer learned by two models trained on 50000 examples. Left: no regularization; right: LIMR with $\sigma=0.5$, $\lambda = 0.7$. LIMR learns filters that resemble those learned by a regularized autoencoder. }
\label{fig:filters1}
\end{figure}

We additionally visualize the filters of the first layer learned by two models in Figure \ref{fig:filters1}. The left panel corresponds to $\lambda = 0$ where no regularization is used; the right panel corresponds to LIMR with $\sigma=0.5$ and $\lambda=0.7$. LIMR learns sharp filters that mostly correspond to pen strokes, which are very similar to those learned by a regularized autoencoder (eg., \cite{cae}). 

\begin{figure}
\centering
\includegraphics[scale=0.4]{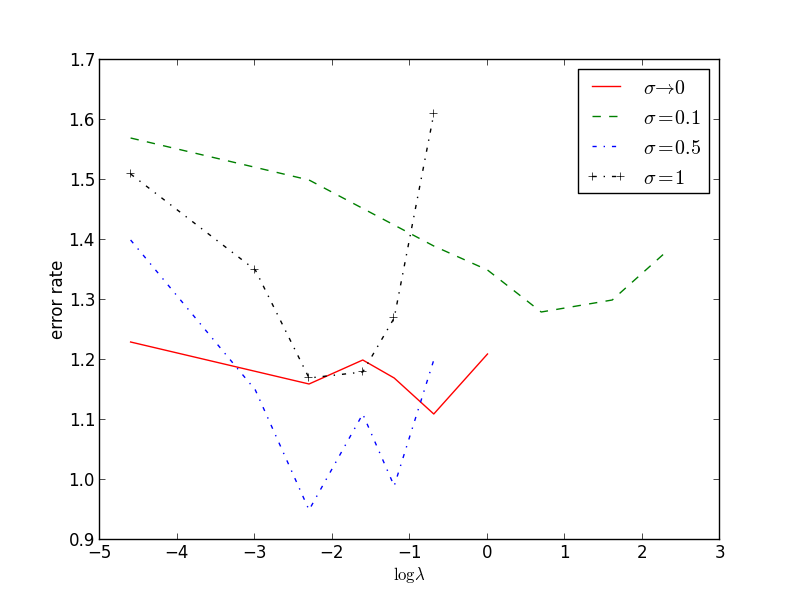}
\caption{Test error rate of using different $\sigma$ values for LAMR under different regularization strengths. X axis is $\log \lambda$, Y axis is test error rate.}
\label{fig:sigma}
\end{figure}

To investigate the property of the stochastic approximation, we have also tried using different values for $\sigma$ from the set $\{\to 0, 0.1, 0.5, 1\}$ to demonstrate its effect on the regularization quality. Here we denote $\to 0$ as using the explicit expression of gradients \cite{dcn} as regularization, as it corresponds to the case where the variance of noise approximates zero. We show the test error rate for different $\sigma$s for LAMR in Figure \ref{fig:sigma}. We see that all the three values of $\sigma$ are able to help reduce the error rate by varying the regularization strength $\lambda$, and $\sigma = 0.5$ achieves the lowest test error rate among the three. In particular, note that taking $\sigma = 0.5$ actually outperforms directly using the gradients as regularization explicitly.
This confirms our speculation that a relatively large $\sigma$ allows the stochastic approximation to explore areas that are further from the observed data points; and by minimizing the difference between the loss of the clean input and the noisy input, the curvature of the target function is further encouraged to be flat.

To compare with the existing methods, we also train this three layer neural network with LAMR and LIMR on 60000 examples by combining the training and validation set and report the test error rate in Table \ref{tb:fc60k}. LAMR sets the state-of-the art result of $0.74$ on this permutation invariant data set (which does not consider the 2D structure of image); LIMR also achieves a result that is close to the previous state-of-the-art.
As a direct comparison, we also try applying feature noising with the same type of noise as used in LAMR and LIMR, which is outperformed by both. Note that we did not try to optimize the architecture of the network at all. In fact, our models have the fewest number of parameters among all the competitors, so it is still possible to reduce the error rate by trying different model sizes.

\begin{table}[h]
\centering
\begin{tabular}{l | l}
Method & Test error rate\\
\hline
$8192 \times 8192$ NN + Dropout \cite{dropout} & 0.95 \\
$500 \times 500 \times 2000$ DBN + Dropout finetuning \cite{dropout} & 0.92 \\
$500 \times 500 \times 2000$ DBM + Dropout finetuning \cite{dropout} & 0.79 \\
$2000 \times 2000$ Contractive Autoenconder \cite{caeh} & 1.04 \\
$2000 \times 2000$ Manifold Tangent Classifier \cite{mtc} & 0.81 \\
$1600 \times 1600$ Maxout NN + Adversarial Training \cite{harness} & 0.782 \\
\hline
$500 \times 500 \times 500$ NN + Feature Noising ($\sigma=0.5$) & 1.06 \\
$500 \times 500 \times 500$ NN + LAMR ($\sigma=0.5, \lambda=0.1$) & \bf 0.74 \\
$500 \times 500 \times 500$ NN + LIMR ($\sigma=0.5, \lambda=0.7$) & 0.83
\end{tabular}
\caption{When trained on 60000 labeled data with the parameters selected from the validation set, LAMR achieves the state-of-the-art test error rate on this permutation invariant version of MNIST; LIMR also achieves a result close to the state-of-the-art.}
\label{tb:fc60k}
\end{table}

We also consider applying our regularizer to CNNs. CNNs are particularly suitable for modeling image data due to the sharing of convolutional filters across different locations within an image. We use an architecture consisting of a convolutional layer of shape $200 \times 9 \times 9$, a max pooling layer of shape $2 \times 2$, another convolutional layer of shape $200 \times 3 \times 3$, another max pooling layer of shape $2 \times 2$, a fully connected layer of size $500$ and finally a linear output layer with $10$ units together with squared hinge loss on top. As in the fully connected network, we also use ReLU as activation function, a $\sigma$ value of $0.5$ and a $\lambda$ value of $0.1$. We train the model with the LAMR, and report the results in Table \ref{tb:conv}. We see that an unregularized CNN already has very low error rates, LAMR is still able to reduce it further. When training on the full 60000 labeled dataset, we achieve a test error that is comparable with the state-of-the-art result in \cite{maxout}.

\begin{table}[h]
\centering
\begin{tabular}{l |l | l | l |l }
&50K, $\lambda = 0$ & 50K, $\lambda = 0.1$ & 60K, $\lambda=0.1$ & 60K, Maxout Network + dropout \\
\hline
validation error rate& 0.76 & 0.52 & - & - \\
test error rate& 0.61 & 0.54 & 0.48 & \bf 0.45
\end{tabular}
\caption{Error rates of CNN trained with LAMR, compared with state of the art model \cite{maxout}. }
\label{tb:conv}
\end{table}

\subsection{Semi-supervised learning with LIMR}
\begin{figure}[t]
\centering
\includegraphics[scale=0.45]{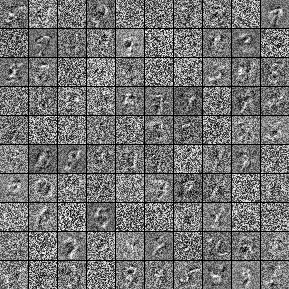}
\includegraphics[scale=0.45]{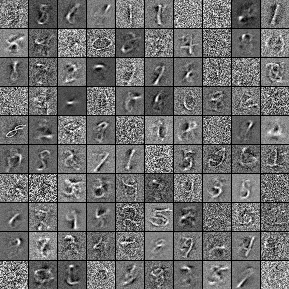}
\par
\vspace{.05cm}
\includegraphics[scale=0.45]{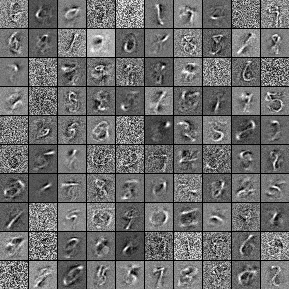}
\includegraphics[scale=0.45]{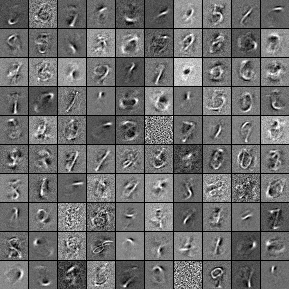}

\caption{Filters learned with different numbers of labels. Number of labels from top left to bottom right: 100, 600, 1000, 3000. We observe a strong correlation between the filter qualities and the number of labels, hence test error rates.}
\label{fig:semi}
\end{figure}
We now consider the semi-supervised learning setting where we only use a small portion of the labels from the training set. We use the same fully connected three layer network as in Section \ref{sect:fc}. We follow the experiment settings as in \cite{mtc,deepsemi}, where the 50000 training set is further split into one labeled set and one unlabeled set. We use the validation set the select $\lambda$ and $\beta$, and found that $\lambda = \beta = 15$ works well for all the four settings. 

\begin{table}[t]
\centering
\begin{tabular}{l | l l l l l}
\# labels & NN & NN + LIMR & CAE & MTC & \cite{deepsemi} \\
\hline
100 &32.61 & 26.04  & 13.47 & 12.03 & \bf 3.33\\
600 & 13.14 & 5.16  & 6.3 & 5.13 & \bf 2.59\\
1000 & 11.15 & 3.03 & 4.77  & 3.64 & \bf 2.40\\
3000 & 6.91 & \bf 1.88  & 3.22 & 2.57  & 2.18
\end{tabular}
\caption{Test error rates with different label numbers. Note that with LIMR we are consistently able to improve the performance of the neural net with unlabeled data. We also achieve the state-of-the-art result on the 3000-labels subtask.}
\label{tb:semi}
\end{table}

\begin{figure}[t]
\centering
\includegraphics[scale=0.5]{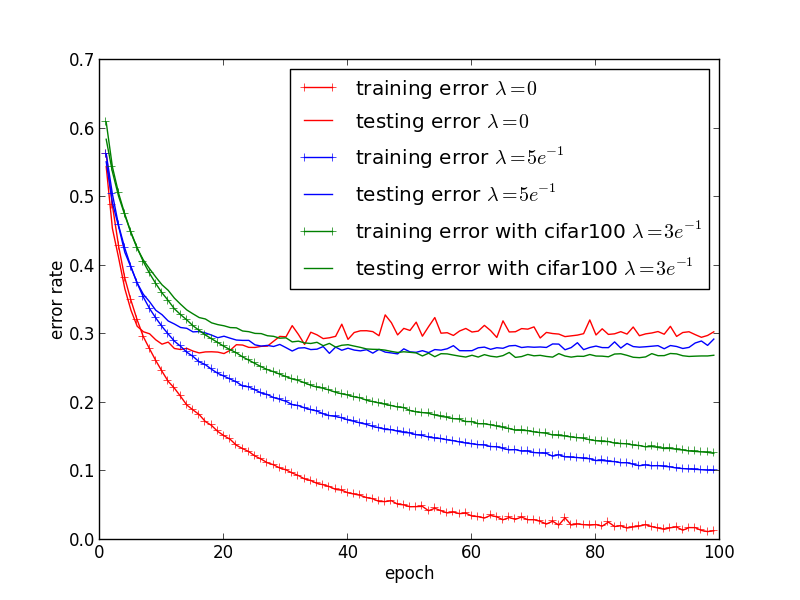}
\includegraphics[scale=0.5]{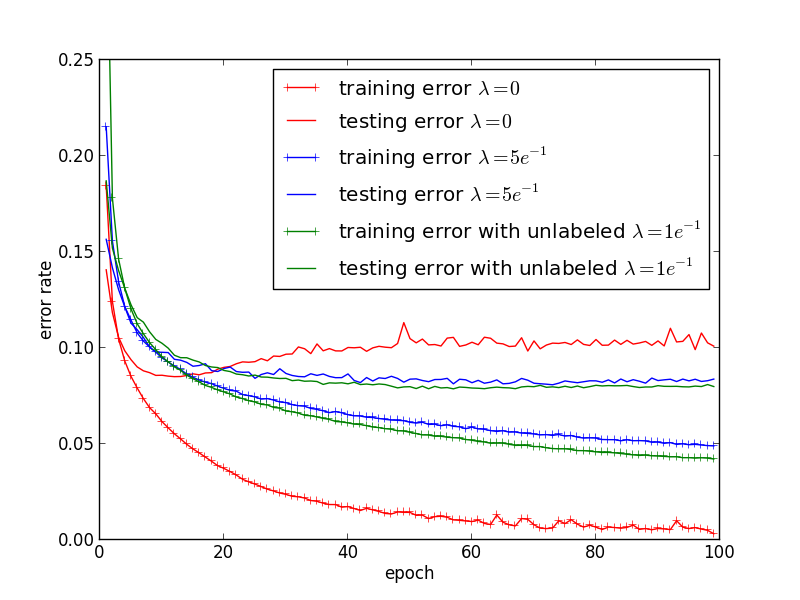}
\caption{Training and testing error curves of CIFAR10 (top) and SVHN (bottom).}
\label{fig:cifar10svhn}
\end{figure}

We report the results in Table \ref{tb:semi}. We see that LIMR consistently boosts the performance of the classifier with the aid of unlabeled data. Compared with the state-of-the-art models, except for the case with only 100 labels, LIMR is comparable or better than pretraining based semi-supervised learning CAE and MTC. The deep generative model proposed by \cite{deepsemi} works the best with very few labels in general, but it is beat by LIMR in the case of 3000 labels. An error rate of $1.88$ also sets the new state-of-the-art result on this sub-task.

As a qualitative justification, we also visualize the first layer filters learned with different label numbers in Figure \ref{fig:semi}. We see that even with a few labels, LIMR is still able to learn sharp filters. There is also a clear correlation between the visual quality of the filters and the number of labels. Especially, the filters learned with 3000 labels are almost indistinguishable from those learned with all the 50000 labels in Figure \ref{fig:filters1}. This strongly indicates the power of semi-supervised learning, as well as that LIMR is able to take advantage of both labels and unlabeled data simultaneously and learn meaningful representations and classifiers at the same time.

We also consider two more challenging benchmarks, CIFAR10 and SVHN. CIFAR10 consists of 60000 $32 \times 32 \times 3$ images in $10$ object categories, with 50000 examples as training set and 10000 as testing set. The subset of the SVHN dataset we use consists of images of the size also $32 \times 32 \times 3$, with 
70000 for training, 26000 for testing and 100000 unlabeled examples. As CIFAR10 does not come with an unlabeled set, we use the entire CIFAR100 dataset for this purpose. Note that although CIFAR10 and CIFAR100 have input images of the same size, they have two different label sets. For both the two benchmarks, we use a CNN with architecture $64 \times 5 \times 5$ $64 \times 3 \times 3$ $64 \times 3 \times$, with $2 \times 2$ max pooling following each convolution layer. We plot the training and testing error curves in Figure \ref{fig:cifar10svhn} under three settings: no regularization, supervised LIMR regularization and semi-supervised LIMR regularization. We see that for both the two datasets, applying LIMR significantly reduces overfitting (larger training error and lower testing error). When incorporated with unlabeled data, LIMR is able to further reduce the testing error. Interestingly, this works despite that CIFAR10 and CIFAR100 have different label sets. 

\section{Conclusion}
We have proposed two regularizers that make use of the data manifold to guide the learning of a discriminative neural network. By encouraging the flatness of either the loss function or the prediction scores along the data manifold, we are able to significantly improve a DNN's generalization ability. Moreover, our label independent manifold regularization allows one to incorporate unlabeled data directly with the supervised learning task and effectively performs semi-supervised learning. We have validated our proposed regularizers on the MNIST, CIFAR10 and SVHN datasets and demonstrated their effectiveness.

\bibliography{iclr2016_conference}

\begin{thebibliography}{19}
\providecommand{\natexlab}[1]{#1}
\providecommand{\url}[1]{\texttt{#1}}
\expandafter\ifx\csname urlstyle\endcsname\relax
  \providecommand{\doi}[1]{doi: #1}\else
  \providecommand{\doi}{doi: \begingroup \urlstyle{rm}\Url}\fi

\bibitem[Bengio et~al.(2007)Bengio, Lamblin, Popovici, Larochelle,
  et~al.]{greedy}
Bengio, Yoshua, Lamblin, Pascal, Popovici, Dan, Larochelle, Hugo, et~al.
\newblock Greedy layer-wise training of deep networks.
\newblock \emph{Advances in neural information processing systems},
  19:\penalty0 153, 2007.

\bibitem[Bishop(1995)]{bishop}
Bishop, Chris~M.
\newblock Training with noise is equivalent to tikhonov regularization.
\newblock \emph{Neural computation}, 7\penalty0 (1):\penalty0 108--116, 1995.

\bibitem[Glorot \& Bengio(2010)Glorot and Bengio]{xavier}
Glorot, Xavier and Bengio, Yoshua.
\newblock Understanding the difficulty of training deep feedforward neural
  networks.
\newblock In \emph{International conference on artificial intelligence and
  statistics}, pp.\  249--256, 2010.

\bibitem[Goodfellow et~al.(2013)Goodfellow, Warde-farley, Mirza, Courville, and
  Bengio]{maxout}
Goodfellow, Ian, Warde-farley, David, Mirza, Mehdi, Courville, Aaron, and
  Bengio, Yoshua.
\newblock Maxout networks.
\newblock In \emph{Proceedings of the 30th International Conference on Machine
  Learning (ICML-13)}, pp.\  1319--1327, 2013.

\bibitem[Goodfellow et~al.(2015)Goodfellow, Shlens, and Szegedy]{harness}
Goodfellow, Ian~J, Shlens, Jonathon, and Szegedy, Christian.
\newblock Explaining and harnessing adversarial examples.
\newblock \emph{ICLR}, 2015.

\bibitem[Gu \& Rigazio(2014)Gu and Rigazio]{dcn}
Gu, Shixiang and Rigazio, Luca.
\newblock Towards deep neural network architectures robust to adversarial
  examples.
\newblock \emph{arXiv preprint arXiv:1412.5068}, 2014.

\bibitem[Hinton \& Salakhutdinov(2006)Hinton and Salakhutdinov]{rbm}
Hinton, Geoffrey~E and Salakhutdinov, Ruslan~R.
\newblock Reducing the dimensionality of data with neural networks.
\newblock \emph{Science}, 313\penalty0 (5786):\penalty0 504--507, 2006.

\bibitem[Kingma et~al.(2014)Kingma, Mohamed, Rezende, and Welling]{deepsemi}
Kingma, Diederik~P, Mohamed, Shakir, Rezende, Danilo~Jimenez, and Welling, Max.
\newblock Semi-supervised learning with deep generative models.
\newblock In \emph{Advances in Neural Information Processing Systems}, pp.\
  3581--3589, 2014.

\bibitem[Konda et~al.(2015)Konda, Bouthillier, Memisevic, and
  Vincent]{dropout_augment}
Konda, Kishore, Bouthillier, Xavier, Memisevic, Roland, and Vincent, Pascal.
\newblock Dropout as data augmentation.
\newblock \emph{arXiv preprint arXiv:1506.08700}, 2015.

\bibitem[Rifai et~al.(2011{\natexlab{a}})Rifai, Dauphin, Vincent, Bengio, and
  Muller]{mtc}
Rifai, Salah, Dauphin, Yann~N, Vincent, Pascal, Bengio, Yoshua, and Muller,
  Xavier.
\newblock The manifold tangent classifier.
\newblock In \emph{Advances in Neural Information Processing Systems}, pp.\
  2294--2302, 2011{\natexlab{a}}.

\bibitem[Rifai et~al.(2011{\natexlab{b}})Rifai, Mesnil, Vincent, Muller,
  Bengio, Dauphin, and Glorot]{caeh}
Rifai, Salah, Mesnil, Gr{\'e}goire, Vincent, Pascal, Muller, Xavier, Bengio,
  Yoshua, Dauphin, Yann, and Glorot, Xavier.
\newblock Higher order contractive auto-encoder.
\newblock In \emph{Machine Learning and Knowledge Discovery in Databases}, pp.\
   645--660. Springer, 2011{\natexlab{b}}.

\bibitem[Rifai et~al.(2011{\natexlab{c}})Rifai, Vincent, Muller, Glorot, and
  Bengio]{cae}
Rifai, Salah, Vincent, Pascal, Muller, Xavier, Glorot, Xavier, and Bengio,
  Yoshua.
\newblock Contractive auto-encoders: Explicit invariance during feature
  extraction.
\newblock In \emph{Proceedings of the 28th International Conference on Machine
  Learning (ICML-11)}, pp.\  833--840, 2011{\natexlab{c}}.

\bibitem[Simard et~al.(2003)Simard, Steinkraus, and Platt]{augment}
Simard, Patrice~Y, Steinkraus, Dave, and Platt, John~C.
\newblock Best practices for convolutional neural networks applied to visual
  document analysis.
\newblock In \emph{null}, pp.\  958. IEEE, 2003.

\bibitem[Srivastava et~al.(2014)Srivastava, Hinton, Krizhevsky, Sutskever, and
  Salakhutdinov]{dropout}
Srivastava, Nitish, Hinton, Geoffrey, Krizhevsky, Alex, Sutskever, Ilya, and
  Salakhutdinov, Ruslan.
\newblock Dropout: A simple way to prevent neural networks from overfitting.
\newblock \emph{The Journal of Machine Learning Research}, 15\penalty0
  (1):\penalty0 1929--1958, 2014.

\bibitem[Szegedy et~al.(2014)Szegedy, Zaremba, Sutskever, Bruna, Erhan,
  Goodfellow, and Fergus]{fooled}
Szegedy, Christian, Zaremba, Wojciech, Sutskever, Ilya, Bruna, Joan, Erhan,
  Dumitru, Goodfellow, Ian, and Fergus, Rob.
\newblock Intriguing properties of neural networks.
\newblock In \emph{International Conference on Learning Representations}, 2014.
\newblock URL \url{http://arxiv.org/abs/1312.6199}.

\bibitem[Tang(2013)]{dlsvm}
Tang, Yichuan.
\newblock Deep learning using linear support vector machines.
\newblock \emph{arXiv preprint arXiv:1306.0239}, 2013.

\bibitem[Vincent et~al.(2010)Vincent, Larochelle, Lajoie, Bengio, and
  Manzagol]{dae}
Vincent, Pascal, Larochelle, Hugo, Lajoie, Isabelle, Bengio, Yoshua, and
  Manzagol, Pierre-Antoine.
\newblock Stacked denoising autoencoders: Learning useful representations in a
  deep network with a local denoising criterion.
\newblock \emph{The Journal of Machine Learning Research}, 11:\penalty0
  3371--3408, 2010.

\bibitem[Wager et~al.(2013)Wager, Wang, and Liang]{lrdrop}
Wager, Stefan, Wang, Sida, and Liang, Percy~S.
\newblock Dropout training as adaptive regularization.
\newblock In \emph{Advances in Neural Information Processing Systems}, pp.\
  351--359, 2013.

\bibitem[Zeiler(2012)]{adadelta}
Zeiler, Matthew~D.
\newblock Adadelta: An adaptive learning rate method.
\newblock \emph{arXiv preprint arXiv:1212.5701}, 2012.

\end{thebibliography}
\bibliographystyle{iclr2016_conference}

\end{document}